\documentclass{article}
\usepackage{spconf,amsmath,graphicx}
\usepackage{subfigure}
\usepackage{epstopdf}
\AppendGraphicsExtensions{.tif}

\title{Defect Detection Approaches Based on Simulated Reference Image}
\name{Nati Ofir, Yotam Ben Shoshan, Ran Badanes and Boris Sherman}

\address{Applied Materials Israel}
\begin{document}
	%\ninept
	%
	\maketitle

	\begin{abstract}
	 This work is addressing the problem of defect anomaly detection based on a clean reference image. Specifically, we focus on SEM semiconductor defects in addition to several natural image anomalies. There are well-known methods to create a simulation of an artificial reference image by its defect specimen. In this work, we introduce several applications for this capability, that the simulated reference is beneficial for improving their results. Among these defect detection methods are classic computer vision applied on difference-image, supervised deep-learning (DL) based on human labels, and unsupervised DL which is trained on feature-level patterns of normal reference images. We show in this study how to incorporate correctly the simulated reference image for these defect and anomaly detection applications. As our experiment demonstrates, simulated reference achieves higher performance than the real reference of an image of a defect and anomaly. This advantage of simulated reference occurs mainly due to the less noise and geometric variations together with better alignment and registration to the original defect background.
	\end{abstract}
	
	
\section{Introduction} \label{sec:intro}

\begin{figure}[tbh]
	\centering
	\includegraphics[width=120px]{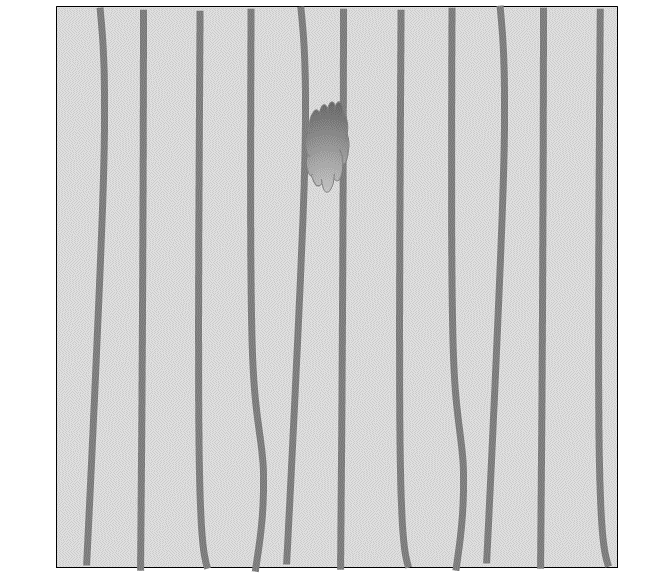}
	\includegraphics[width=120px]{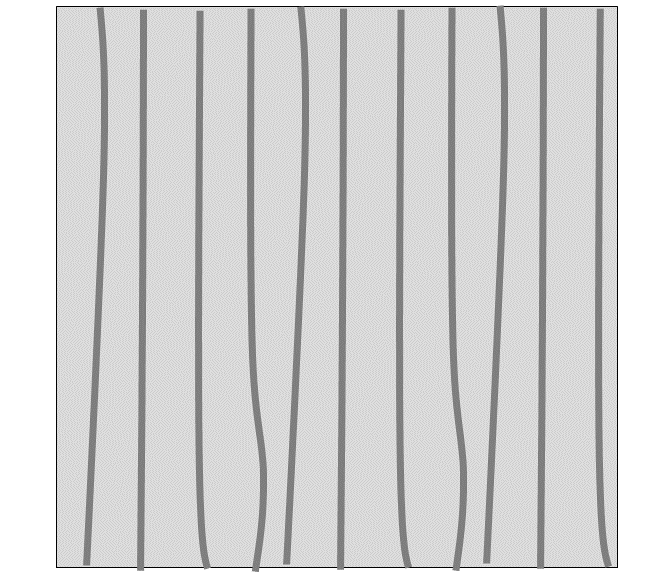}
	\caption{Example of a simulation of Scanning-Electron-Microscopy (SEM) image with anomaly and its corresponding clean reference image.}
	\label{fig:example}
\end{figure}

In this manuscript, we introduce two approaches to simulating the clean reference of a defect image and several methods that utilize this reference. The different schemes that can benefit from a simulated reference can be classic, supervised, or unsupervised, we will cover each such solution in this paper. See Figure \ref{fig:example} for an illustration of a defect in the Scanning-Electron-Microscopy (SEM) image and its clean reference image. Denote the defect or possibly clean image by $I(x)$, and its reference defect-free image by $R(x)$. In this work, we will utilize generative Convolutional-Neural-Network (CNN) to produce a simulated reference image such that $CNN(I(x)) = \hat{R}(x)$. A candidate method to generate such simulated reference is by inpainting as described in \cite{zavrtanik2021reconstruction}. A second generative model to clean defects and anomalies in an image is by Variational-Auto-Encoder (VAE) like its very-deep version \cite{child2020very}.

\begin{figure}[tbh]
	\centering
	\includegraphics[width=120px]{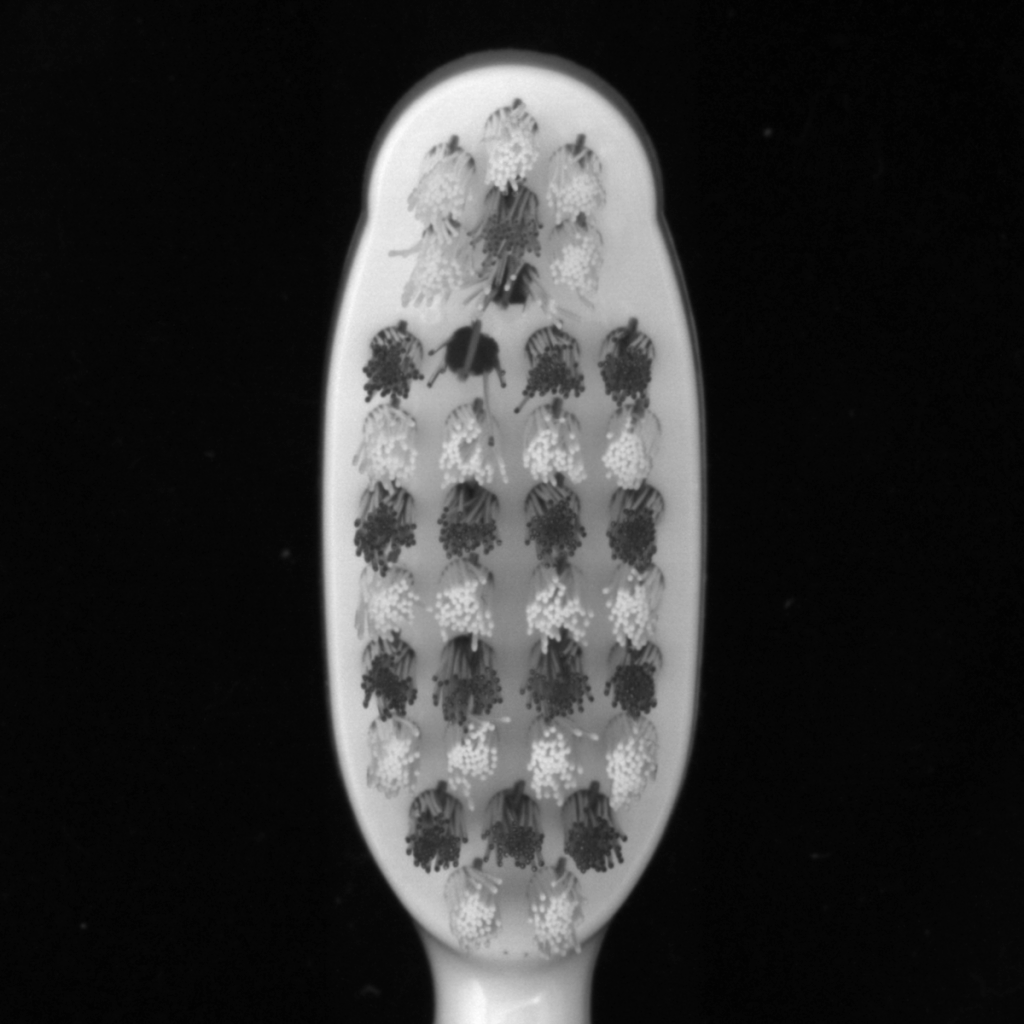}~
	\includegraphics[width=120px]{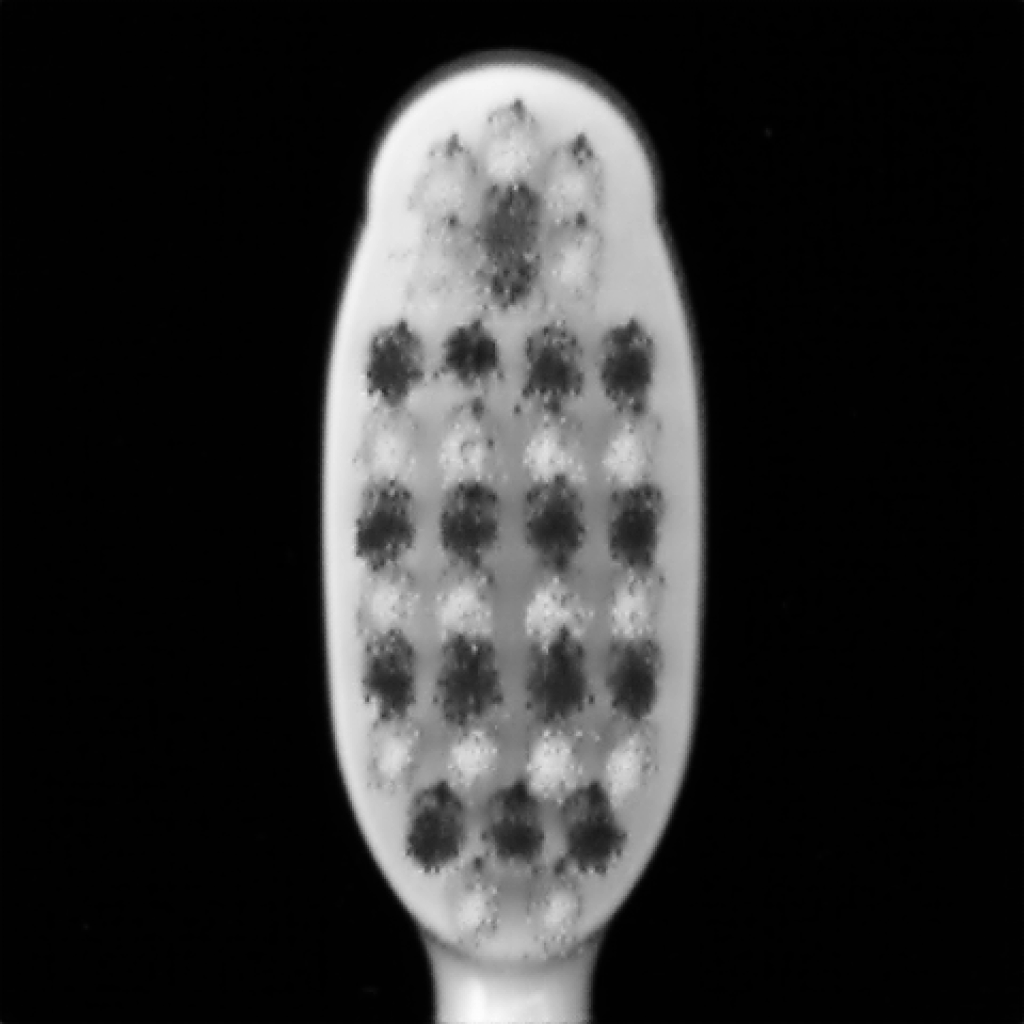}
	\caption{Left: A real image of a toothbrush with anomaly from the dataset of \cite{bergmann2021mvtec}. Right: A simulated reference image where the anomaly is eliminated.}
	\label{fig:real_example}
\end{figure}

In Figure \ref{fig:real_example}, can be seen a toothbrush defective image from the MVTec anomaly detection dataset \cite{bergmann2021mvtec}. This figure shows the simulated reference $\hat{R}(x)$ for this specific toothbrush generated from reconstruction by inpainting method \cite{zavrtanik2021reconstruction}. As shown in this Figure, the simulated reference preserves the nominal structure of the toothbrush while changing only the anomaly pixels. This characteristic can be utilized to detect anomalies by different approaches including, classic computer vision, and supervised and unsupervised learning.
 
This paper is organized as follows: In Section \ref{sec:previous} we cover previous work on the topic of simulated reference images and anomaly detection. In Section \ref{sec:simulate} we explain the two main methods for generating defect-free images using deep generative learning. In Section \ref{sec:classic} we introduce how to use simulated reference in classic defect detection. In Section \ref{sec:supervised} and Section \ref{sec:unsupervised} we explain how to incorporate simulated reference in supervised and unsupervised deep learning correspondingly. In Section \ref{sec:experiments} we prove that simulated reference in contributing to defect detection both quantitatively and qualitatively, and Finally we conclude this manuscript in Section \ref{sec:conclusions}.

\section{Previous Work} \label{sec:previous}

Defect and anomaly detection is a fundamental problem of computer vision with a plethora of related works. Early methods relied on classic machine learning \cite{lane1997application}, image processing techniques like Gabor filters \cite{kumar2002defect} and Fourier analysis \cite{chan2000fabric}. Advanced methods used Deep-Learning (DL) for defect detection in fabric \cite{cseker2016fabric} and in semiconductor manufacturing \cite{yang2022semiconductor}. Recent work introduced an approach for supervised and unsupervised detection of defects in SEM images using anomaly learning \cite{9898035}.

Generative anomaly detection using simulated reference is a relevant problem studied in the last decade. Recent methods utilized Generative-Adversarial-Network (GAN) for unsupervised anomaly detection \cite{schlegl2017unsupervised}. Wang et al. used VAE to apply generative anomaly detection \cite{wang2018generative}. Reconstruction by inpainting was introduced to create a simulated reference from defective image \cite{zavrtanik2021reconstruction}.

Anomaly and defect detection can be carried out by representation approaches. Padim \cite{defard2021padim} model the patch distribution to find anomalies. PatchCore \cite{roth2022towards} learns a memory bank of nominal images to find anomalies by Approximate-Nearest-Neighbour (ANN) technique \cite{arya1998optimal}. This work introduces an approach to apply defect detection by a combination of generative and representation methods, our experiment shows that using this method improves overall accuracy over using the real reference image.

\section{Simulating Reference Image} \label{sec:simulate}

\subsection{Inpainting}

\begin{figure}[tbh]
	\centering
	\includegraphics[width=250px]{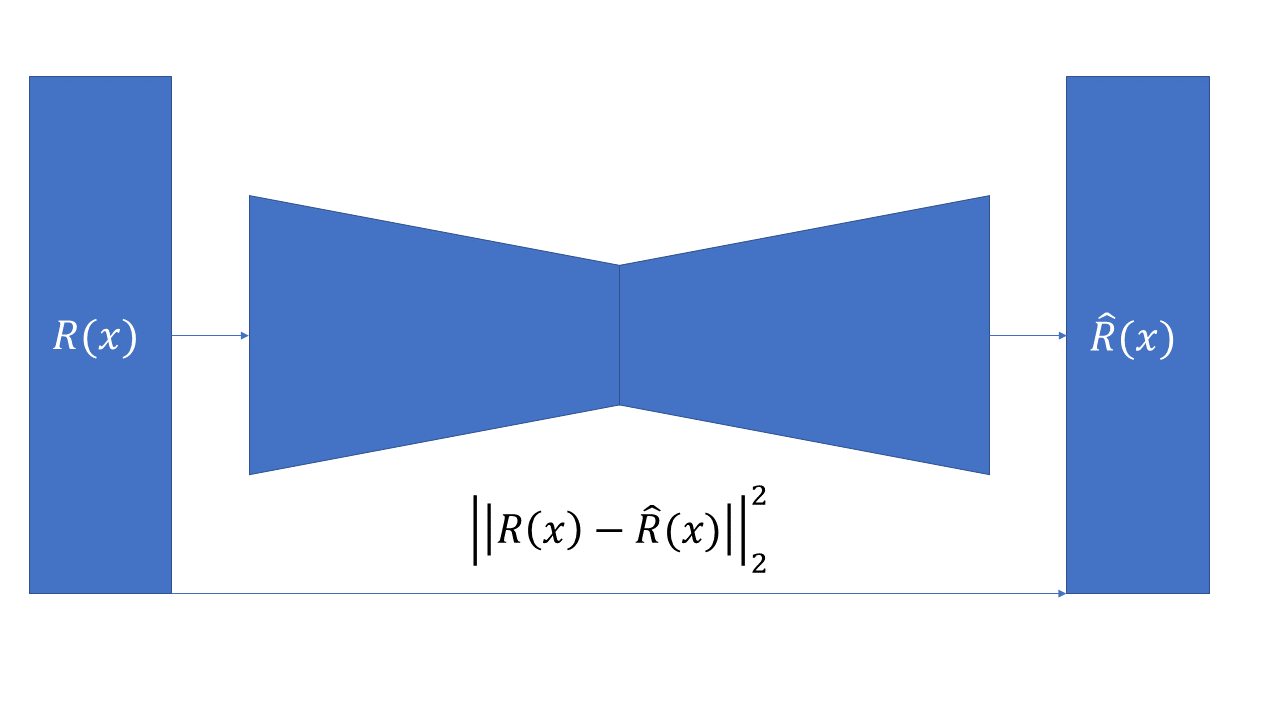}
	\caption{CNN architecture of training and generating simulative reference image. In the training phase, we supply as input the real reference $R(x)$ while in test time we use the defect candidate image $I(x)$.}
	\label{fig:generative}
\end{figure}

An approach for defect erasing for the creation of simulated reference images can use image inpainting techniques like was introduced in \cite{zavrtanik2021reconstruction}. Given a reference image $R(x)$, we mask it with an inpainting mask $B(x)$, and compute a forward pass in a generative neural network: $CNN(R(x)\cdot B(x))$. In the training time, we would constraint that the output of the neural network will reconstruct the clean reference image by a Mean-Square-Error loss:
\begin{equation}
    ||CNN(R(x)\cdot B(x)) - R(x)||_2^2.
\end{equation}
After training the neural network with a real reference image dataset, we will compute the simulated reference on a defect image $I(x)$ such that $\hat{R}(x) = CNN(I(x))$. This simulated reference will clean the defect and preserve the background patterns. See Figure \ref{fig:generative} for an illustration of the architecture for generating the simulated reference image.

\subsection{Variational Auto Encoder}

A different method for generalization of a clean reference image, given a defect image is by Variationl-Auto-Encoder (VAE) such as the very-deep variant of \cite{child2020very}. The idea behind using VAEs is to train them again with a dataset of real reference images. The normalization of the defect image in test time will occur due to the normal distribution bottleneck of the VAE. We enforce the VAE to contain a normal distribution latent space by training it with Evidence lower bound (ELBO) \cite{yang2017understanding} containing the Kullback-Libler (KL) Divergence loss \cite{joyce2011kullback}. The VAE is eliminating defects in test time and contributes to computing the simulated reference images.

\section{Classic Defect Detection} \label{sec:classic}

The simulated reference image can be utilized by classic defect detection that compares the defect image to its reference by simple difference with post-processing. Given a defect image $I(x)$, and its real reference image $R(X)$, we can compute the absolute difference image $D(x) = |I(x)-R(x)|$. By applying an advanced image and post-processing we can find the different pixel locations in $D(x)$. Given a simulated reference computed by a generative CNN $\hat{R}(x)$ we compute the simulated difference image:
\begin{equation}
    \hat{D(x)} = |I(x)-\hat{R}(x)|.
\end{equation}
Our experiments in Section \ref{sec:experiments} show that detecting defects by the simulated reference yield higher accuracy, this is due to the fact that it contains less noise, and process variation, and is better aligned to the defect image $I(x)$.

\section{Defect Detection by Supervised Deep Learning} \label{sec:supervised}

The simulated reference can be used as an input to a Convolutional Neural Network (CNN) that produces defect detection and is trained based on human labels. The annotation process, which is significantly manual, can be partially automated by the creation of reference simulation and brings the first suggestions for annotation by the classic or unsupervised approaches.

We detect defects by a U-Net CNN \cite{du2020medical} with a Residual-Network (Resnet) backbone \cite{wu2019wider}. Denote by $U$ our defect detection network. A forward pass of $U$ is done over a pair of defects and reference to compute a segmentation map $S(x) = U(I(x), R(x)$. Given a simulated reference image, we can derive the simulated segmentation map: 
\begin{equation}
    \hat{S(x)} = U(I(x), \hat{R}(x)).    
\end{equation}
We train the network by a Cross-Entropy loss \cite{zhang2018generalized}, with balancing between foreground and background. Afterward, we test the segmentation network in a similar approach and compute the relevant metrics. Our experiments in Section \ref{sec:experiments} proved that training and testing this unsupervised CNN with a simulated reference achieves better results than using the real reference images.

\section{Simulated Reference for Unsupervised Anomaly Detection} \label{sec:unsupervised}

A dataset of clean reference images can be used for unsupervised anomaly detection as done by PatchCore algorithm \cite{roth2022towards}, which was implemented in anomaly detection library \cite{akcay2022anomalib}. PatchCore-like is one of a group of representation-based anomaly detection. It detects defects and anomalies by comparing their representation to a memory bank of nominal descriptors. Our proposed method replaces this memory bank with representations of simulated reference images. Thus, whether the train set contains defects or nominal images, we clean them by our CNN and compute the simulated memory bank of clean image features.

\begin{figure}[tbh]
	\centering
	\includegraphics[width=250px]{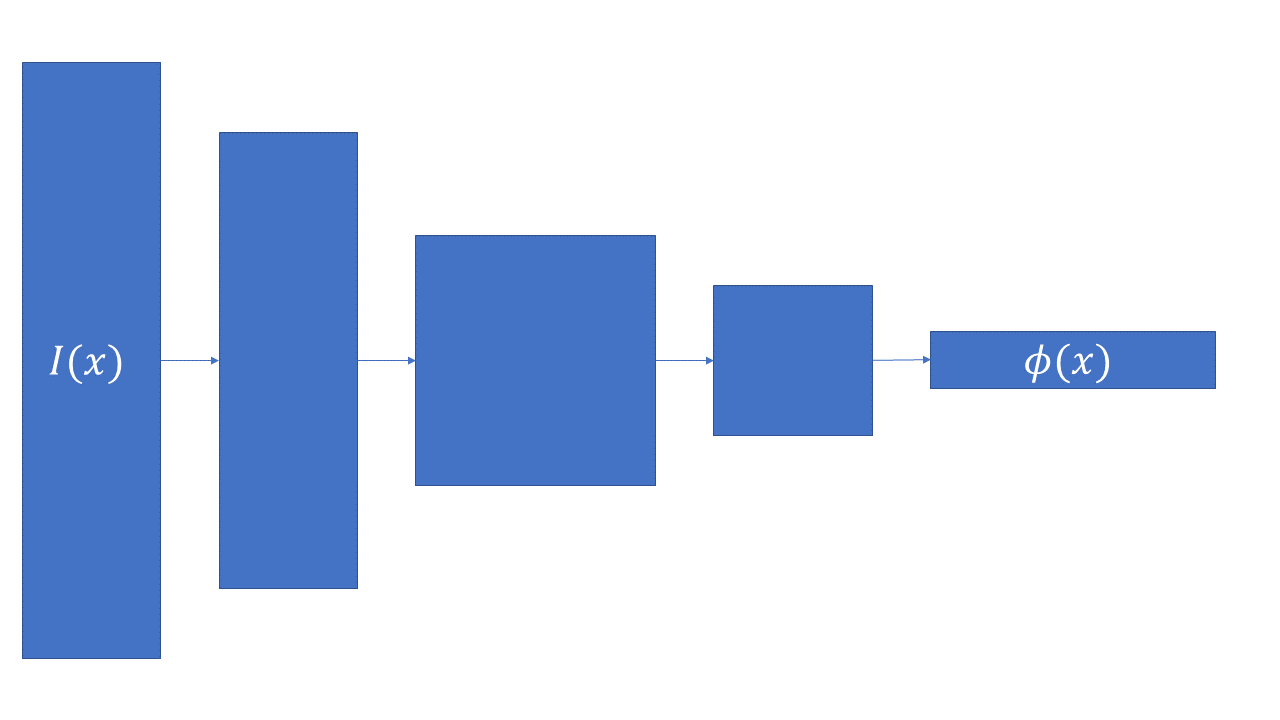}
	\caption{CNN architecture of generating the representation feature vector $\phi(x)$ given an input image $I(x)$.}
	\label{fig:representation}
\end{figure}

Given a dataset of real reference images $R_1(x),..., R_(n)$, we compute its memory bank of feature using pre-trained Resent backbone: $M = \phi_1(x),..., \phi_k(x)$. See Figure \ref{fig:representation} for an illustration of the architecture for generating the representation feature vector $\phi(x)$.

Then, for a defect image $I(x)$ we compute the ANN distance to M: $dist(I(x), M) = ANN(I(x), M)$. By that metric, we produce anomaly maps. Given simulated reference images $\hat{R_1}(x),..., \hat{R_n}(x)$, we repeat the same procedure to compute the simulated memory bank $\hat{M}$. Our experiments in Section \ref{sec:experiments} show that the simulated approach better classifies the defect images as defective or nominal. Moreover, its trainset is robust to cases where a defective pattern accidentally appears in the nominal image trainset.

\section{Experiments} \label{sec:experiments}

 We tested our approaches for simulating reference images on classic defect detection using difference-image, supervised DL using human labels, and on the unsupervised method of PatchCore \cite{roth2022towards}. We used two datasets for evaluating our approaches: the internal SEM semiconductor dataset and the anomaly detection dataset, MVTec \cite{bergmann2021mvtec}.

\begin{figure}[tbh]
	\centering
    \includegraphics[width=120px]{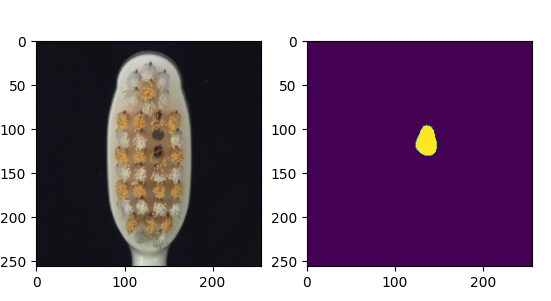} \\
    \includegraphics[width=120px]{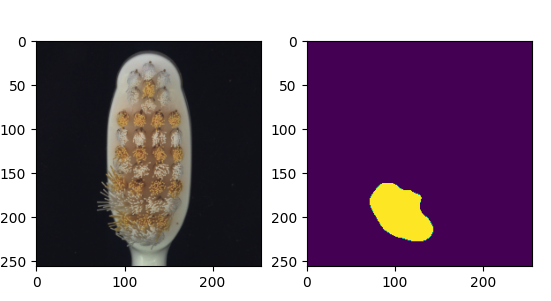} \\
    \includegraphics[width=120px]{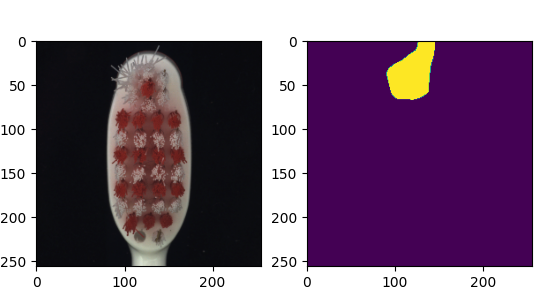} \\
    \includegraphics[width=120px]{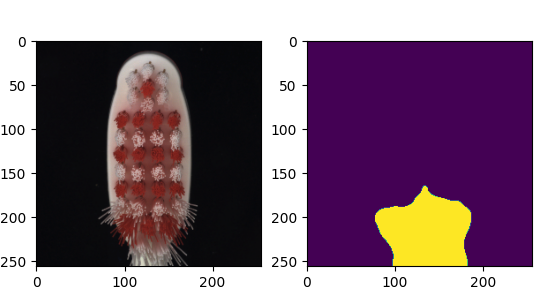} \\
    \includegraphics[width=120px]{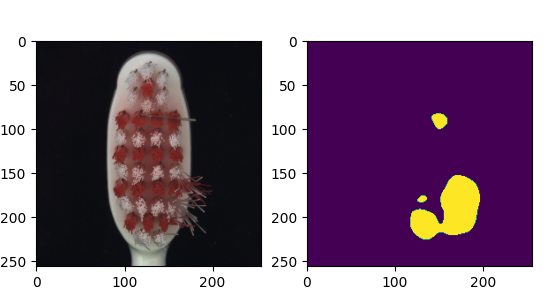} \\
    \includegraphics[width=120px]{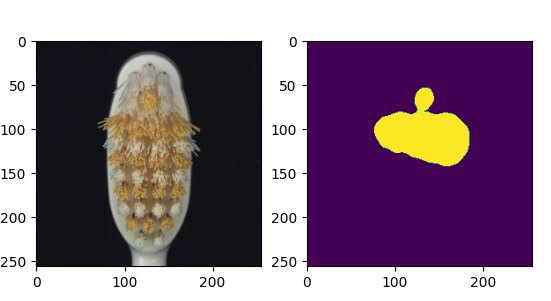} \\
	\caption{Prediction masks on anomalies on the toothbrush dataset of \cite{bergmann2021mvtec} based on simulated reference trainset applied with the Patch Core method \cite{roth2022towards}.}
	\label{fig:results}
\end{figure}
 
\begin{table}[tbh]
	\centering%
	\begin{tabular}{|l|c|c|c|}
		\hline
		Classic/Layer & A (CR/FR) & B (CR/FR) & C (CR/FR)\\ \hline\hline
        Simulated Reference& 0.94/0.97 & 0.75/0.97 & 0.85/0.93 \\ \hline
        Real Reference&  0.90/0.92 & 0.80/0.97 & 0.82/0.82 \\ \hline
		\end{tabular}%
	\caption{Results of our classic solution based on difference image maps of simulated reference versus real reference. The result format is the Capture Rate (CR) and Filter Rate (FR) of our dataset containing both true defects and clean background images.}
	\label{table:classic}
\end{table}

\begin{table}[tbh]
	\centering%
	\begin{tabular}{|l|c|c|c|}
		\hline
		Supervised/Layer & A (CR/FR) & B (CR/FR) & C (CR/FR) \\ \hline\hline
        Simulated Referencce& 0.99/0.99 & 0.98/0.75 &  0.95/0.97 \\ \hline
        Real Reference& 0.98/0.99 & 0.99/0.92 &  0.92/0.97 \\ \hline
		\end{tabular}%
	\caption{Results of our supervised deep-learning (DL) solution based on human labels of simulated reference versus real reference. The result format is the Capture Rate (CR) and Filter Rate (FR) of our dataset containing both true defects and clean background images.}
	\label{table:supervised}
\end{table}

\begin{table}[tbh]
	\centering%
	\begin{tabular}{|l|c|c|}
		\hline
		Unsupervised & Classification F-score \\ \hline\hline
        Simulated Reference & 0.94 \\ \hline
        Real Reference &  0.91\\ \hline
		%B real & 0.91 & 0.93 \\ \hline
        %B simulated & 0.99 & 0.5 \\ \hline
	\end{tabular}%
	\caption{Results of Patch Core \cite{roth2022towards} algorithm for anomaly detection based on real reference image dataset versus simulated dataset on and SEM semiconductor dataset from 'C' layer of SEM semiconductor.}
	\label{table:patchcore}
\end{table}

\begin{table}[tbh]
	\centering%
	\begin{tabular}{|l|c|c|}
		\hline
		Unsupervised & Classification F-score\\ \hline\hline
        Simulated Reference & 1.0 \\ \hline
        Real Reference & 0.98 \\ \hline
		\end{tabular}%
	\caption{Results of Patch Core \cite{roth2022towards} algorithm for anomaly detection based on real reference image dataset versus simulated dataset the toothbrush dataset of \cite{bergmann2021mvtec}}
	\label{table:patchcoreNatural}
\end{table}

Table \ref{sec:classic} show our capture rate (CR) which is a recall of defective, and filter rate (FR), that is Recall of non-defective patterns, on the classic approach for defect detection using a clean reference image. Note that $Recall = \frac{tp}{tp+fn}$, such that $tp = true-positive$ and $fn = false-negative$. This table compares the performance on three semiconductor layers for real reference images versus simulated ones. It can be seen that in most cases, simulated references yield better CR/FR. This happens due to less process variation, less noise, and better alignment to the defective pattern.

Similar results for unsupervised DL, using human labels are described in Table \ref{sec:supervised}. The advantage of the simulated reference as an input versus its corresponding real reference is emphasized by the overall CR/FR in most of the semiconductor layers.

Table \ref{table:patchcore} shows the F-score of classification of an image as defective or nominal, on the anomaly detection dataset of MVTec \cite{bergmann2021mvtec}. In this case, the training of the PatchCore \cite{roth2022towards} is done on a dataset of clean reference images. This table compares the usage of a simulated trainset versus a real reference dataset. Also in this experiment, usage of the simulated reference images improves the overall accuracy due to less noise and geometric variations. The images in this table are semiconductors from a 'C' layer of the previous experiments.

Similar results are shown in Table \ref{table:patchcoreNatural}, for natural images of toothbrush from the anomaly detection dataset \cite{bergmann2021mvtec}. This proves that our approach to improving representation anomaly detection with generative methods works on various types of images. It can be extended to other representation approaches that are using a clean dataset as a reference for defect detection.

An illustration of our end-to-end anomaly detection, using simulated reference is shown in Figure \ref{fig:results}. For every image of an anomaly defect, we produce an anomaly segmentation map. 

Overall, all these experiments prove quantitatively and qualitatively that utilizing simulated reference images improves all types of defect detection algorithms, including, classic, supervised, and unsupervised DL.

\section{Conclusion} \label{sec:conclusions}
We introduced several usages for the simulation of reference images, that benefited from their advantages. Among them, classic defect detection based on difference-image, supervised DL by human labels, and unsupervised DL carried out by training on a clean reference dataset. In all these applications, our experiments demonstrated the advantages of using the simulated reference over the real one. This fact is due to its low level of noise and geometric variations and better alignment for registration of the defect image background. In addition to the improvement of measurements, this enables building systems for anomaly detection that are not based on reference grabbing and therefore efficient and optimized.

	{\small
		\bibliographystyle{ieee}
		\bibliography{egbib}
	}
	
\end{document}